\documentclass[runningheads]{llncs}
\usepackage[T1]{fontenc}
\usepackage{graphicx}
\graphicspath{{../}{./}}
\usepackage{amsmath,amssymb}
\usepackage{xcolor}
\usepackage{cite}
\usepackage{marvosym}
\usepackage[
  colorlinks=true,
  linkcolor=blue,
  citecolor=blue,
  urlcolor=blue
]{hyperref}
\begin{document}
\title{Closed-Loop LLM Discovery of \\Non-Standard Channel Priors in Vision Models}
\titlerunning{LLM-Guided Channel NAS}
\author{Tolgay Atinc Uzun\textsuperscript{(\Letter)}, 
Dmitry Ignatov, and 
Radu Timofte}
\authorrunning{T. A. Uzun et al.}
\institute{Computer Vision Lab, CAIDAS \& IFI, University of W\"urzburg, Germany
\email{t.atincuzun@gmail.com}}
\maketitle              
\begin{abstract}
Channel-configuration search—the optimization of layer specifications such as channel widths in deep neural networks—presents a combinatorial challenge constrained by tensor-shape compatibility and computational budgets.
We investigate whether Large Language Models (LLMs) can support Neural Architecture Search (NAS) by reasoning over architectural code structures in ways that complement traditional search heuristics.
In this paper, we apply an LLM-driven NAS framework to channel-configuration search, formulating the task as conditional code generation in which the LLM refines architectural specifications using performance feedback.
To address data scarcity, we generate a corpus of valid, shape-consistent architectures through Abstract Syntax Tree (AST) mutations.
While these mutated networks are not necessarily optimized for performance, they provide structural examples that help the LLM learn executable architectural patterns and relate channel configurations to model performance.
Experimental results on CIFAR-100 show that the closed-loop LLM improves upon the initial AST-generated architecture population under the same proxy-evaluation protocol.
Our analysis further shows that the generated architectures reflect domain-specific design patterns, including non-standard channel widths and late-stage expansion, highlighting the potential of language-driven design for code-level NAS. The code and prompts are publicly available at \href{https://github.com/ABrain-One/NN-GPT}{https://github.com/ABrain-One/NN-GPT}, and the generated deep neural networks are published at \href{https://github.com/ABrain-One/NN-Dataset}{https://github.com/ABrain-One/NN-Dataset} under model names with the prefix \texttt{ast-dimension-}.

\end{abstract}
\section{Introduction}
Neural architecture design is typically formulated as a discrete optimization problem and solved using reinforcement learning~\cite{zoph2016neural}, evolutionary algorithms, or differentiable relaxations~\cite{liu2018darts}.
These methods generally treat the model definition as a fixed graph and do not utilize the semantic structure of executable source code.
Large Language Models (LLMs) provide an alternative route for Neural Architecture Search (NAS): they can modify neural-network programs directly, using pre-trained knowledge of code syntax and logical flow to support iterative architectural refinement~\cite{kochnev2025optuna,Rupani2025llm,Gado2025llm,ABrain.NN-RAG}.

A primary challenge in applying LLMs to NAS is the lack of domain-specific training data.
Standard repositories do not contain sufficient variations of executable architectures to train a model on structural design.
This study addresses this limitation through a programmatic bootstrapping mechanism using Abstract Syntax Tree (AST) manipulation, which generates a large corpus of syntactically valid and tensor-consistent network variants for initialization.

This research studies the use of an LLM as a NAS optimizer in the concrete setting of channel configuration search.
Modifying layer widths requires maintaining consistency across coupled network components such as residual connections, making the setting suitable for testing whether closed-loop code generation can improve architecture definitions under strict structural constraints.
Training on the generated corpus and subsequent performance feedback enables the model to iteratively refine channel configurations to improve accuracy.

\section{Related Work}

\subsection{Neural Architecture Search}
Conventional NAS uses reinforcement learning~\cite{zoph2016neural} and differentiable search~\cite{liu2018darts} to automate model design across vision backbones and efficient transformer language models~\cite{so2021evolving}, but usually relies on restrictive predefined supernets or search spaces.
Generative architecture methods broaden this view, yet valid neural-network code remains discrete and tightly constrained.
RL-based generators typically select layers and connections from a fixed vocabulary, limiting exploration to known subspaces; our method instead treats architecture design as open-ended token-level code modification, enabling patterns not explicitly encoded in the search space.

\subsection{Automated Channel Pruning and Model Compression}
Channel-configuration optimization is central to model compression.
He et al.~\cite{he2018amc} use a deep deterministic policy gradient agent to sample per-layer compression ratios under accuracy, FLOP, and latency constraints.
Where this treats channel counts as continuous RL actions, our method treats them as code tokens, allowing the LLM to exploit syntax and program logic rather than numerical rewards alone.
Network slimming~\cite{liu2017slimming} prunes channels through L1-regularized batch-normalization scales, removing near-zero channels after training; in contrast, we optimize the architecture definition before training by generating efficient source-code configurations.
MetaPruning~\cite{liu2019metapruning} trains a hypernetwork to predict weights for pruned networks, enabling many channel configurations to be evaluated without full retraining.
Our work predicts the code defining those channels, framing the problem as language modeling rather than weight prediction.
Direct channel-number search has also been studied in one-shot settings such as AutoSlim~\cite{wang2020channelnet}.

\subsection{LLM-Driven Optimization and Discovery}
LLM-based evolutionary search is rapidly emerging.
EvoPrompting~\cite{xu2023evoprompting} uses LLMs as mutation operators, replacing random bit flips or subtree exchange with code-aware recombination and function optimization.
FunSearch~\cite{romera2024funsearch} discovers mathematical algorithms through sandboxed code evolution, parallel populations, and feedback from the highest-scoring executable programs, surpassing human baselines in domains such as bin packing.
FunBO~\cite{aglietti2024funbo} similarly uses LLMs to discover acquisition functions for Bayesian optimization.

Prompt-based NAS methods, including GPT-4-enhanced NAS~\cite{wang2023llmnas} and GPT-NAS~\cite{zhang2023gptnas}, ask LLMs to propose layers or connections directly.
GPT-NAS encodes architectures as token sequences and fine-tunes a transformer to predict high-performing strings, while LeMo-NADe~\cite{rahman2024lemonade} uses expert-system validation for latency- and power-aware edge-device discovery.
However, prompt-only methods rely on pre-trained knowledge for validity; under constraints such as residual channel consistency, this can produce invalid candidates and waste evaluation budget.
We instead fine-tune on a programmatically generated, task-guided corpus of valid code.

\subsection{Code Generation and Validation}
Generated-code validity is central to LLM-based search.
LLMatic~\cite{nasir2023llmatic} combines LLMs with MAP-Elites quality-diversity optimization, preserving trade-offs such as model size versus accuracy rather than converging to a single model.
Self-repair methods~\cite{olausson2023selfrepair} use compiler feedback to fix LLM errors.
NNGPT~\cite{kochnev2025nngpt} proposes a closed-loop AutoML engine that fine-tunes the LLM on successful discoveries, while Optuna vs Code Llama~\cite{kochnev2025optuna} shows that fine-tuned LLMs can predict optimal hyperparameters zero-shot, challenging Bayesian optimization.
Related NNGPT studies further highlight the promise of LLM-driven network design~\cite{ABrain.NN-RAG,ABrain.NN-Captioning_2025,ABrain.NNGPT-Fractal,ABrain.Transform,ABrain.Architect,ABrain.Prompt}.

Synthetic code data is also gaining traction.
\textit{Textbooks are All You Need}~\cite{gunasekar2023textbooks} introduces Phi-1, trained on textbook-quality synthetic Python code, showing that curated synthetic data can yield strong reasoning.
\textit{Code Alpaca}~\cite{chaudhary2023codealpaca} applies instruction tuning to synthetic code instructions, aligning with evidence that LLMs learn useful program-generation behavior from code corpora~\cite{radford2019language,chen2021evaluating}.
These findings support our use of AST-generated architectures as high-quality synthetic data.
Rather than using ASTs post hoc for repair or relying on zero-shot prediction, we use them a priori to teach validity, letting the LLM focus on pattern matching and optimization rather than basic syntax.

\section{Methodology}
\label{sec:Methodology}
We propose a generative approach to NAS where an LLM acts as an intelligent optimizer.
Rather than relying only on heuristic mutations, we condition the LLM to synthesize neural network architectures that target improved performance.
The pipeline consists of three phases: (1) AST-based bootstrapping to initialize the knowledge base, (2) conditional generative optimization in which the LLM proposes improved architectures from metric targets, and (3) an iterative fine-tuning loop that refines the model using high-performing code structures.

\subsection{Problem Formulation: Conditional Code Generation}
We frame the channel configuration search as a conditional code generation task.
Let $\mathcal{M}$ be the space of valid neural network source codes.
Given a baseline model $m_{base} \in \mathcal{M}$ with performance $y_{base}$, and a desired target performance $y_{target} > y_{base}$, the LLM must synthesize a new model $m_{new}$:
\begin{equation}
    m_{new} = \text{LLM}(m_{base}, y_{base}, y_{target}).
\end{equation}
The goal is for the actual performance of $m_{new}$ to approximate or exceed $y_{target}$.
This formulation explicitly conditions the generation on the performance metric, requiring the LLM to reason about the relationship between code structure (channel widths) and model capacity. Unlike standard code completion, which optimizes for likelihood $P(code)$, our objective is to optimize for the conditional probability of improvement: $P(m_{new} | m_{base}, y_{new} > y_{base})$.

\begin{figure}
    \centering
    \includegraphics[width=1\linewidth]{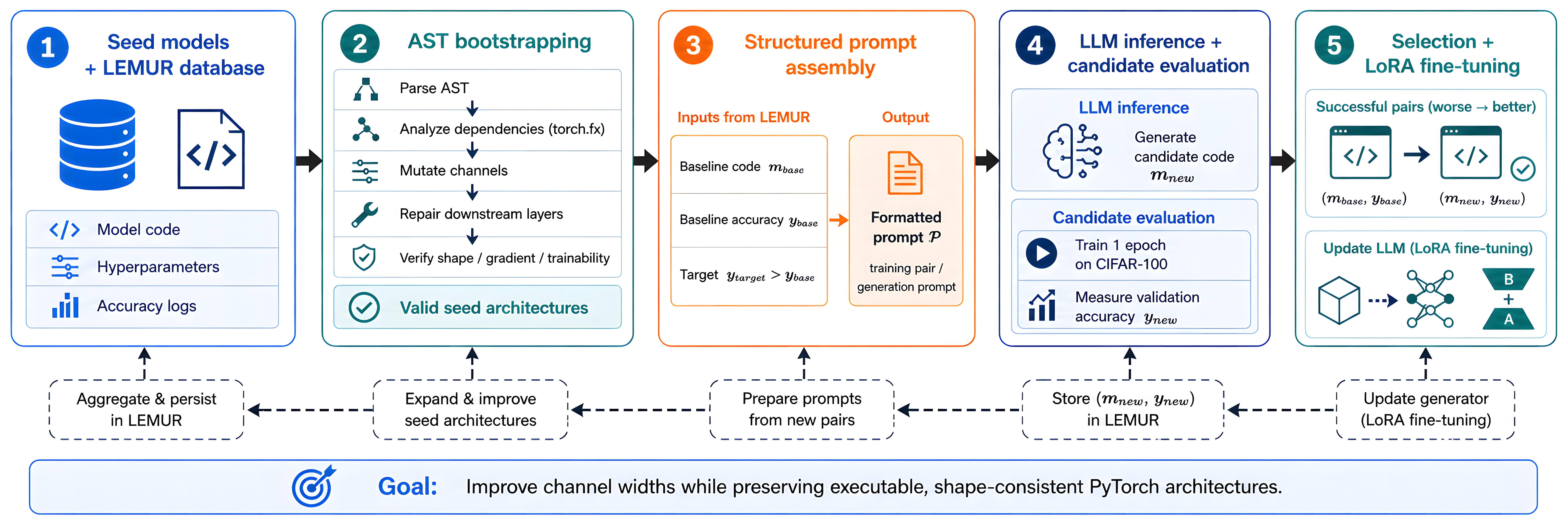}
    \caption{Overview of our closed‑loop search pipeline.  The system first retrieves seed architectures and associated training logs from the LEMUR database.  These examples are mutated by an AST‑based bootstrapping engine that rewrites layer widths while preserving tensor consistency, populating an initial pool of valid candidates.  During the main loop, the baseline code and its accuracy are fed into a structured prompt that conditions the LLM to generate a new architecture and hyperparameter set.  Each candidate is trained for a single epoch and its performance logged.  Successful generations are added back into the database and used to fine‑tune the LLM, thus closing the feedback loop.}
    \label{fig:go2}
\end{figure}

\subsection{The LEMUR Database}
To support this data-driven process, we utilize the LEMUR dataset of high-capacity and edge-optimized neural networks~\cite{lemur2025,ABrain.LEMUR2,ABrain.NN-Lite}.
LEMUR serves as the persistent memory for our system, storing every generated model as executable PyTorch code alongside its training metadata (hyperparameters, epoch count) and evaluation metrics (validation accuracy).
This unified storage allows us to query for model pairs and construct training datasets for the LLM dynamically.

While LEMUR serves as a comprehensive repository for networks and their metadata, applying it to the specific problem of channel optimization presents a cold-start challenge. To orient the LLM towards this task, we cannot rely on an empty or generic database; we need a rich set of examples that demonstrate the causal link between channel configuration and model performance. Therefore, we systematically populate the database with generated models to serve as the initial training corpus, ensuring the LLM can discern the nuances of the optimization landscape from the onset of fine-tuning.

\subsection{Phase 1: Bootstrapping via AST Mutation}
LLM requires domain-specific examples to understand the task of channel configuration.
To solve the cold starting issue, we employ a programmatic mutation engine solely for the initial epoch.
This engine parses the AST of seed models and applies structural changes using a three-stage pipeline. As a seed model, we chose AlexNet~\cite{krizhevsky2012imagenet}, as it provides a simple starting point for channel configuration search.

\begin{figure}
    \centering
    \includegraphics[width=1\linewidth]{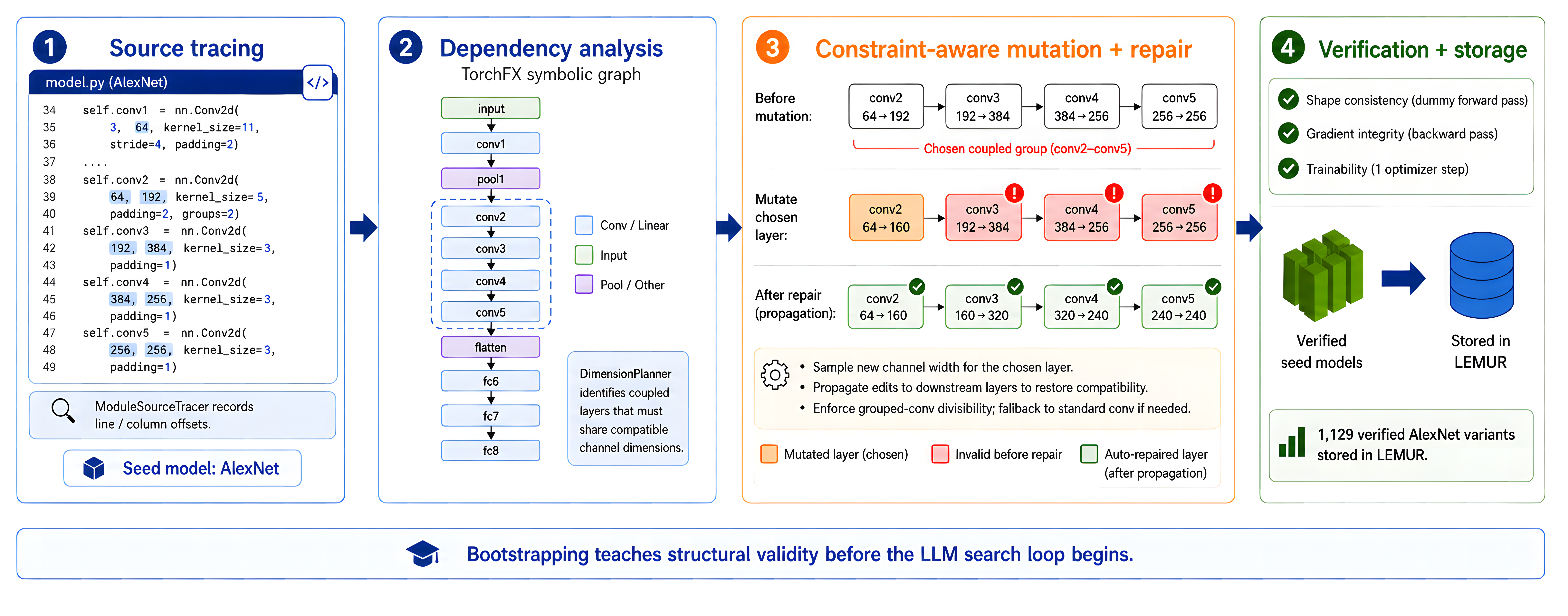}
    \caption{The AST-based bootstrapping pipeline. It illustrates the process of AST parsing (extracting layer definitions and offsets), dependency analysis using the TorchFX symbolic graph, the constraint-aware editing and repairing phase, and the verification protocol using dummy tensor checks to generate valid seed models. Whenever a random mutation is chosen and applied, the network must be corrected. The invalid layers are shown in red. By examining the input and outputs to Conv2D layers, the network is corrected later on.}
    \label{fig:astbootstrapper}
\end{figure}

\subsubsection{1. Source Tracing and Planning}
We implement a \texttt{ModuleSourceTracer} that patches the \texttt{\_\_init\_\_} methods of PyTorch modules (e.g., \texttt{nn.Conv2d}, \texttt{nn.Linear}) to capture their exact line numbers and column offsets during instantiation.
A \texttt{ModelPlanner} then analyzes the model's data flow graph (using \texttt{torch.fx} with a custom \texttt{LeafTracer}) to identify mutation groups, sets of layers that must share dimensions due to downstream dependencies.
This ensures that if a convolutional layer's output width is modified, the corresponding input width of the next layer (and any parallel branches) is updated synchronously.

\subsubsection{2. Constraint-Aware Execution}
The \texttt{CodeMutator} applies the mutation plan directly to the source code AST.
The AST mutation engine relies on Python's \texttt{ast} module for source edits and \texttt{torch.fx} for constraint-aware planning.
While the engine supports various structural modifications including activation swapping, kernel size adjustments, and stride changes, our primary focus in this work is on channel dimension optimization.
Crucially, the system ensures validity by propagating these changes across the entire network, automatically correcting downstream layers to maintain graph consistency.
\begin{itemize}
    \item \textbf{Dimension Mutation:} The \texttt{DimensionPlanner} selects a mutation group and assigns a new channel width from a predefined valid range ($[4, 1025]$). It enforces architectural constraints, such as ensuring that new channel counts remain divisible by the \texttt{groups} parameter in grouped convolutions. If a depthwise convolution's constraint is violated, the system automatically resets it to a standard convolution.
\end{itemize}

\subsubsection{3. Verification Protocol}
Before any mutated model is added to the database, it undergoes a rigorous verification process in the \texttt{Orchestrator}:
\begin{enumerate}
    \item \textbf{Shape Consistency:} A forward pass with synthetic data (e.g., $2 \times 3 \times H \times W$) confirms that tensor shapes remain valid throughout the network and that the final output matches the dataset's class count.
    \item \textbf{Gradient Integrity:} A backward pass ensures the computational graph remains differentiable.
    \item \textbf{Trainability Check:} A single optimizer step is executed to verify that parameters can be updated.
\end{enumerate}
This phase generates the initial population of valid, trainable models. Once the database is populated, the AST engine is retired, and all subsequent generation is driven by the LLM. This process yielded 1,129 verified AlexNet variants that populated the initial LEMUR database.

\subsection{Phase 2: Conditional Generative Optimization}
We construct training samples by querying LEMUR for pairs of models $(m_A, m_B)$ trained on the same dataset, where $m_B$ outperforms $m_A$ ($y_B > y_A$).
This pairing logic is handled by the \texttt{JoinConf} configuration in our data loader, which explicitly filters for pairs where the addon model has a higher metric than the baseline.
Crucially, $m_B$ need not be a direct descendant of $m_A$; the pairing simply provides a trajectory from a lower-performing architecture to a higher-performing one.

\subsubsection{Prompt Engineering}
We design a structured prompt that explicitly conditions the LLM on the metrics.
The input prompt contains:
\begin{itemize}
    \item \textbf{Role:} "You are a machine learning model designer."
    \item \textbf{Task:} "Generate a LEMUR dataset neural network model that increases the '{metric}' metric value to at least {addon\_accuracy}..."
    \item \textbf{Context:} The source code of the baseline model $m_A$ (wrapped in \texttt{<nn>} tags), its hyperparameters (\texttt{<hp>}), and its achieved accuracy $y_A$.
\end{itemize}
The LLM is tasked with generating the full source code for $m_B$ (wrapped in \texttt{<nn>} tags) and its corresponding hyperparameters.
By training on these pairs, the LLM learns to analyze the baseline architecture and synthesize the necessary structural modifications (e.g., widening specific bottlenecks, balancing layer ratios) to achieve the target metric.

\subsection{Phase 3: Iterative Fine-Tuning Loop}
The search proceeds in iterative epochs. In each epoch $t$:
\begin{enumerate}
    \item \textbf{Sampling:} We sample baseline models from the current LEMUR database.
    \item \textbf{Conditional Generation:} We prompt the LLM with these baselines and a \textit{higher} target accuracy (e.g., $y_{target} = y_{base} + \delta$). The LLM synthesizes $N$ new candidate architectures.
    \item \textbf{Evaluation:} The candidates are trained for a fixed number of epochs (proxy training). Their code and resulting accuracies are stored in LEMUR.
    \item \textbf{Fine-Tuning:} We identify successful generations (where the new model actually improved over the baseline) and other high-performing pairs from the updated database. The LLM is fine-tuned on this fresh data.
\end{enumerate}
This loop creates a self-reinforcing cycle. As the database accumulates better models, the LLM is exposed to higher-quality code examples and more ambitious improvement trajectories, progressively refining its ability to design optimal architectures.

\section{Experiments and Results}
\subsection{Experimental Setup}
Fine-tuning of LLMs and training of computer vision models are performed using the AI $\text{Linux}$ docker image $\texttt{abrainone/ai-linux}$\footnote{AI Linux: \url{https://hub.docker.com/r/abrainone/ai-linux}} on NVIDIA GeForce RTX 3090/4090 $24\text{G}$ GPUs of the $\text{Kubernetes}$ cluster and a dedicated workstation.

We evaluate the efficacy of our closed-loop NAS pipeline on the CIFAR-100 dataset, using an AirNet\cite{chee2018airnet}-based vision architecture as the search skeleton. Model performance is assessed by validation accuracy after a single training epoch for each candidate image classification model.
The search space is intentionally restricted to the channel configurations (layer widths) of all convolutional and fully connected layers, constrained by the structural consistency requirements detailed in Section ~\ref{sec:Methodology}.
The optimization process is initialized with a bootstrapping phase (epoch 0) consisting of AST-mutated variants to populate the initial candidate pool.
The search is conducted over a 22 iteration trajectory (epochs 0--21).
Each candidate architecture is evaluated using a proxy metric: validation accuracy after a single training epoch, following a standardized training recipe (batch size of 64 and the AdamW optimizer) and employing advanced data augmentation techniques~\cite{Aboudeshish2025augmentation}; the reported gains should therefore be interpreted within this proxy-evaluation setting.

All generation and fine-tuning are conducted using 
\texttt{OlympicCoder-7B}~\cite{penedo2025olympiccoder}, a 7-billion 
parameter code-oriented LLM with a context window of 16{,}384 tokens.
We employ parameter-efficient fine-tuning using LoRA (Low-Rank Adaptation)~\cite{hu2021lora} with rank $r=32$, alpha $\alpha=32$, and dropout $0.05$.
Adapters are applied to the query, key, value, and output projection matrices (\texttt{q\_proj}, \texttt{k\_proj}, \texttt{v\_proj}, \texttt{o\_proj}).
Training uses the \texttt{paged\_adamw\_8bit} optimizer, following the memory-efficient optimizer setup popularized by QLoRA~\cite{dettmers2023qlora}, with a cosine learning rate scheduler and a learning rate of $10^{-6}$.
Generation uses the Hugging Face text-generation pipeline with stochastic decoding (temperature $0.8$, top-$k$ $70$, top-$p$ $0.9$).

\subsection{Generative Validity Analysis}
The search process generated 220 candidate architectures across the initial 22 epochs.
The strict enforcement of trainability and testability resulted in a validity rate of 9.09\%: 20 candidates satisfied all structural constraints, while 200 failed the verification protocol.
This low validity rate highlights a central difficulty of unconstrained source-code NAS, where small channel edits can violate coupled tensor dependencies across downstream or residual layers.
Invalid candidates are rejected before evaluation and therefore do not affect the reported accuracy statistics; however, they reduce search efficiency and indicate an important direction for future work, such as constrained decoding or repair-guided generation.
Despite the sparsity of valid samples, the successful candidates provided useful signal for improving generated channel configurations.

\subsection{Search Trajectory and Performance Gains}
\begin{figure}[htbp]
    \centering
    \includegraphics[width=0.48\textwidth]{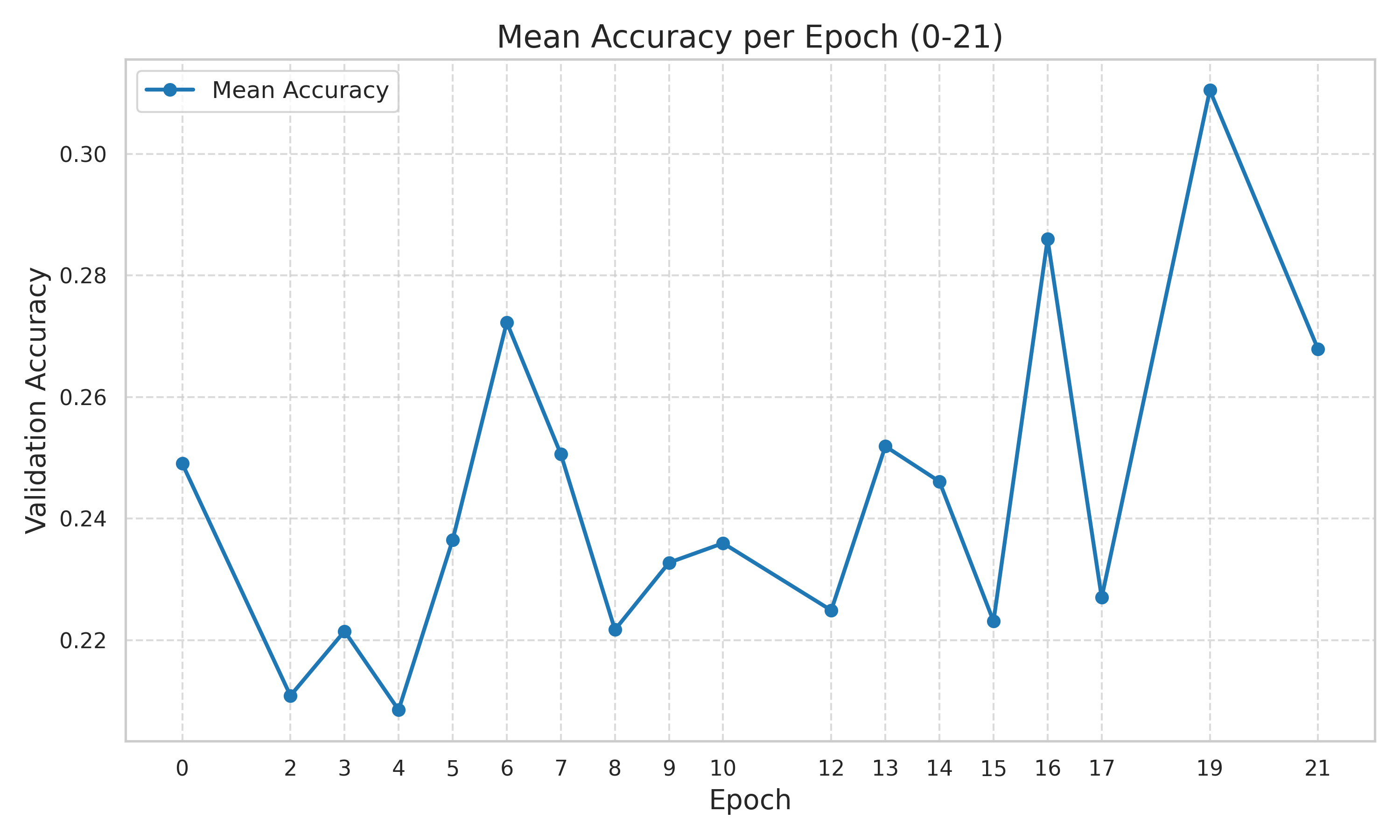}
    \includegraphics[width=0.48\textwidth]{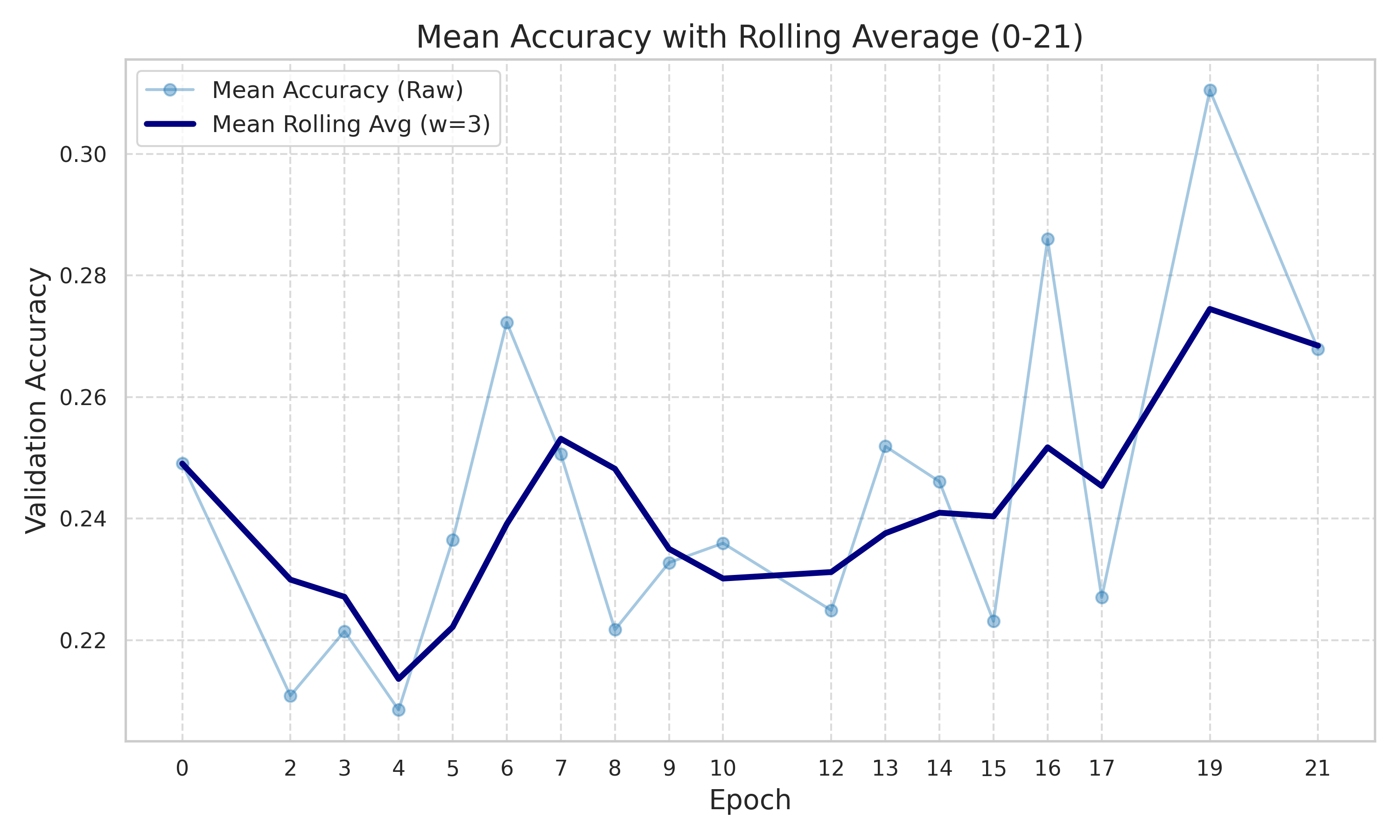}

    \caption{CIFAR-100 validation accuracy (one training epoch) of LLM-generated image classification models across LLM fine-tuning epochs. Left: mean accuracy of all valid models; fluctuations reflect the exploration of diverse configurations. Right: rolling average of validation-set classification accuracy (window size = 3) for image classification models generated by an LLM across fine-tuning epochs. }
    \label{fig:accuracy_metrics}
\end{figure}

\begin{figure}[htbp]
    \centering
    \includegraphics[width=0.48\textwidth]{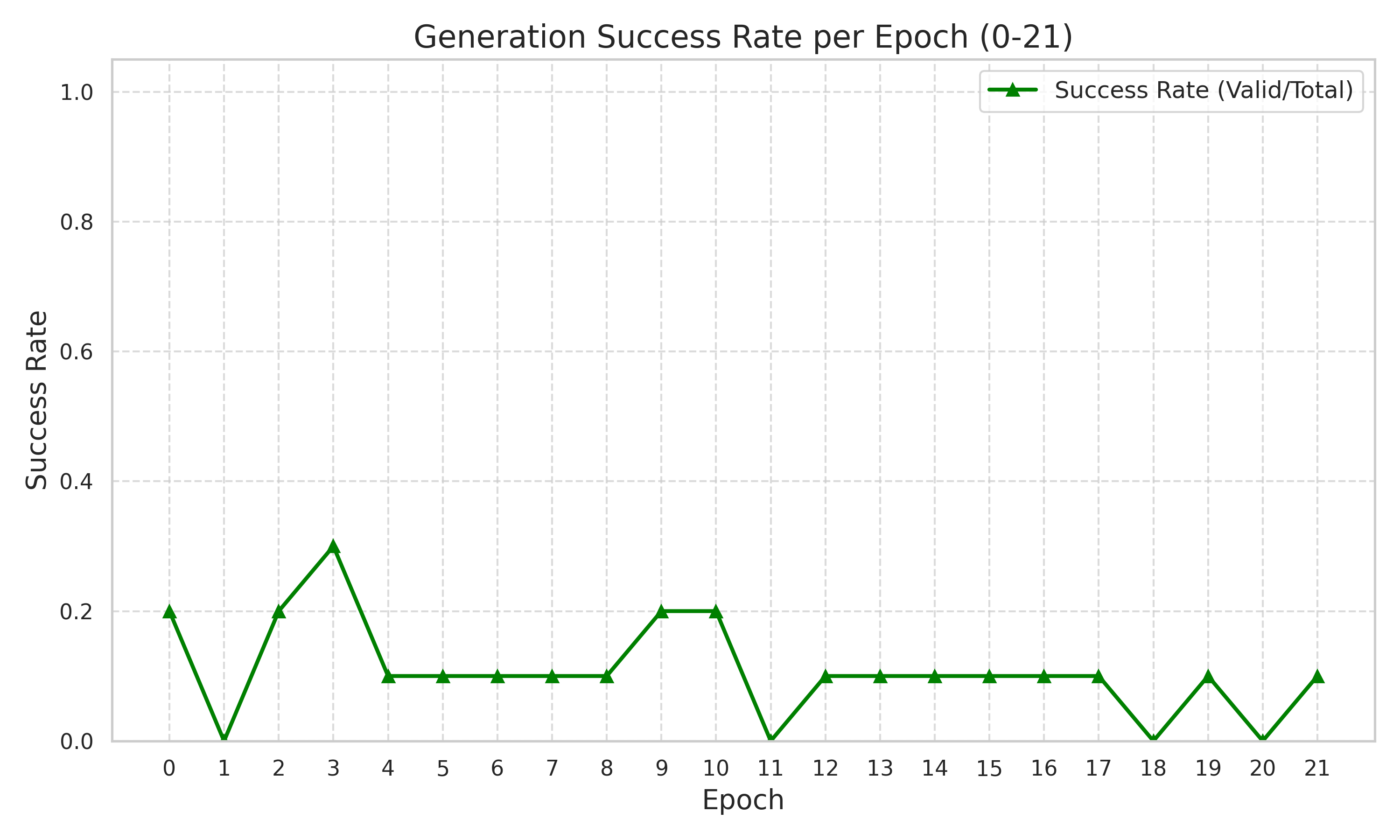}
    \includegraphics[width=0.48\textwidth]{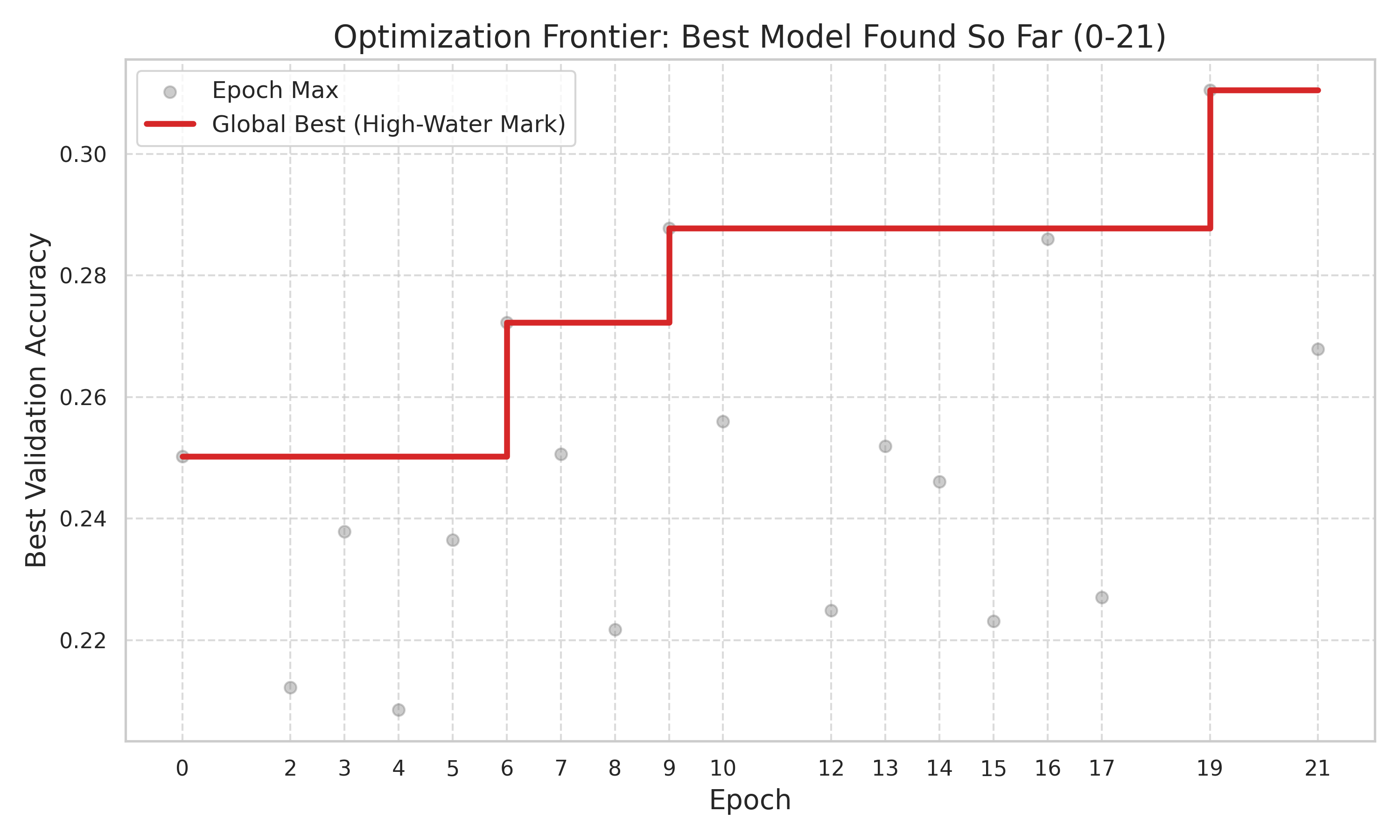}
    \caption{Search dynamics. Left: generation success rate (valid/total), illustrating the difficulty of producing structurally valid code under channel constraints. Right: best-so-far trajectory (high-water mark), where step-function jumps highlight discrete improvements in the generated architectures.}
    \label{fig:success_best}
\end{figure}

The evolution of the model performance is illustrated in Fig.~\ref{fig:accuracy_metrics} and Fig.~\ref{fig:success_best}.
The search process commenced with a baseline accuracy of 0.250 established during the initialization phase.
Through iterative refinement, the LLM identified architectural configurations that exceeded this initial population.
The global maximum accuracy achieved was 0.311 (epoch 19), representing a 24.1\% relative improvement over the best model found in the initial AST-generated distribution under the same proxy-evaluation protocol.

To visualize the optimization process, we present a comprehensive analysis of the search trajectory across 21 epochs.
The raw performance metrics are presented in Fig.~\ref{fig:accuracy_metrics} left, while the plot on the right provides a smoothed view to highlight the underlying optimization signal.
Finally, Fig.~\ref{fig:success_best} details the generation success rate and the cumulative best performance of the system.

\subsection{Statistical Analysis of Improvement}
We employed a multi-faceted statistical analysis to evaluate the search trajectory.

\subsubsection{Peak Performance Trend (Linear Regression)}
We analyzed the progression of the maximum accuracy per epoch to quantify the optimization of the trajectory.
A linear regression analysis on the epoch-wise maximum accuracy reveals a positive slope ($\beta = +0.0019$) with a $p$-value of $0.083$.
While this $p$-value is marginally above the standard $0.05$ threshold, linear regression assumes a constant monotonic trend, whereas discrete NAS trajectories often contain alternating exploration phases and abrupt improvements.
Thus, the positive slope suggests an upward trend, and we complement it with non-parametric tests that make fewer distributional assumptions.

\subsubsection{Population Analysis (T-Test)}
To assess whether the earlier models displayed improvement over epochs, we compared the model populations from the initial exploration phase, the epochs 0-5, against those from the late optimization phase which is taken as epochs 16-21.
First, we analyzed the entire population of valid models.
The mean accuracy increased from $0.226$ in the early phase to $0.273$ in the late phase.
A one-tailed t-test supports this improvement with a $p$-value of $0.0033$.
This shift in the population mean is consistent with the LLM capturing useful regularities between channel allocation and validation accuracy.
Rather than serving only as an unconstrained sampler, the closed-loop model increasingly proposes configurations associated with higher predictive performance among valid candidates.

\subsubsection{Robust Frontier Analysis (Permutation Test)}
Given the limited sample size of the late-stage population ($N=4$), we employed a non-parametric permutation test to evaluate the observed improvement without assuming normality.
We compared the early phase population ($N=9$, $\mu=0.226$) against the late phase population ($N=4$, $\mu=0.273$).
By simulating $100,000$ random permutations of the group labels, we determined that the probability of observing a performance gap of this magnitude ($\Delta = +0.0474$) by chance is \textbf{$p = 0.0077$}.
This result provides distribution-free evidence that later valid candidates are shifted toward higher accuracy, although the small number of valid late-stage samples warrants caution.



\section{Discussion}

\subsection{Efficacy of AST-Based Bootstrapping}
The programmatic initialization of the search space was a determining factor in the system's convergence.
By seeding the repository with 1129 syntactically verified examples, the system decoupled the acquisition of structural constraints from the optimization objective.
This allowed the LLM to leverage the initial corpus as a syntactic prior, focusing its capacity on refining channel configurations for accuracy rather than learning the rules of valid code generation from scratch. Notably, although the seed examples consisted exclusively of AlexNet channel configurations, the LLM successfully extrapolated this structural knowledge to the block-level architecture of AirNet.

\subsection{Prompt Importance Ablation}

To assess the necessity of structured prompting, we conducted a prompt-removal ablation in which the task definition, metric target, and dataset context were removed jointly. This experiment does not isolate individual prompt components, but it tests whether structured conditioning as a whole is required for validity and task alignment. Across 220 attempts, only 9 candidates were executable. The absence of dataset constraints also caused domain drift: 5 of the 9 valid models were inadvertently architected for the CelebA Gender task rather than the target CIFAR-100 dataset. The remaining 4 CIFAR-100 candidates displayed negligible capacity, yielding validation accuracies between 0.0202 and 0.108. These results support the role of explicit conditioning in both performance optimization and constraining the LLM to the correct problem domain.

\subsection{Architectural Priors and Late-Stage Expansion}
To interpret the design patterns acquired during the closed-loop process, we analyzed the correlation between the width of specific layers and the final model accuracy.
The analysis reveals a non-uniform sensitivity to channel capacity, suggesting that the LLM learned to reallocate computational resources rather than simply scaling them uniformly.

\subsubsection{Resource Reallocation}
We observed a strong negative correlation of $\rho = -0.47$ between the width of the second layer and model accuracy, contrasted with a strong positive correlation of $\rho = +0.53$ for the fourth layer.
This pattern is consistent with a topology characterized by late-stage expansion.
While standard architectures often follow a uniform doubling pattern, the high-performing configurations generated by the LLM tend to compress the early feature extraction layers and expand the final representational layer.
The global maximum model exemplifies this with a configuration of $[64, 99, 256, 1536]$.
In this architecture, the second layer is reduced to 99 channels, while the final layer is expanded to 1536 channels.
This suggests that, within this search space, the LLM favored compact early feature extraction followed by a high-dimensional projection space for class separation.

\subsubsection{Non-Standard Channel Widths}
A significant portion, specifically $40.9\%$, of the generated layer widths were not powers of two.
The presence of specific values such as $99$ or $308$ in high-performing models challenges the standard engineering heuristic of aligning dimensions to hardware registers.
The LLM treated channel count as a continuous hyperparameter.
These configurations suggest trade-offs between information bottlenecking and parameter count that are not restricted to the discrete power-of-two grid.

\subsubsection{Parameter Efficiency and Pareto Optimality}
\begin{figure}[htbp]
    \centering
    \includegraphics[width=1.0\textwidth]{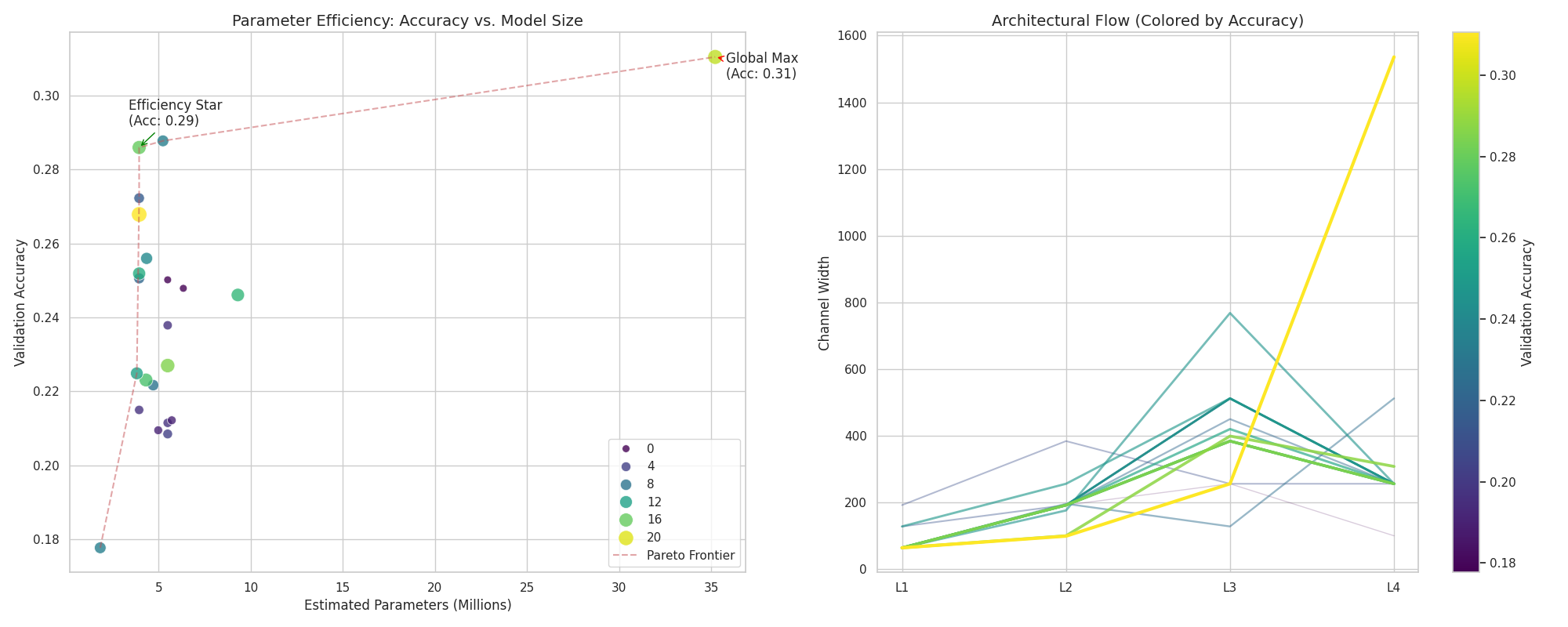}
    \caption{Efficiency and flow analysis. Left: accuracy vs. parameter count. The red dashed line indicates the Pareto frontier of efficient models. The LLM generates models that maximize accuracy for a given parameter budget. The color and the size of the points remark the epoch that the models were generated. Right: architectural flow (parallel coordinates) for all models, colored by accuracy. High-performing models (yellow/green) consistently exhibit a specific trajectory: narrow early layers followed by a massive expansion in the final layer.}
    \label{fig:channel_profiles}
\end{figure}

Beyond raw accuracy, we analyzed the relationship between model size and performance.
Fig.~\ref{fig:channel_profiles} illustrates the accuracy versus parameter landscape.
We observe a distinct Pareto frontier where the LLM generates architectures that achieve high accuracy with fewer parameters than the baseline.
The most efficient model achieves comparable performance to the global maximum but with a reduced parameter footprint, suggesting that the closed-loop process can favor efficient configurations even without an explicit efficiency objective.
The architectural flow analysis shows that these efficient high-performers share common structural characteristics.
They employ a strategy where early layers are kept narrow to save parameters, while the final layers are expanded to maximize representational capacity.

\section{Conclusion}
We introduced a closed-loop neural architecture search framework in which a large language model optimizes channel configurations for vision networks by operating directly on executable source code. By casting channel search as conditional code generation, the approach departs from graph-based or numerical NAS formulations and enables optimization under strict structural constraints. A central contribution is a validity-first bootstrapping mechanism based on AST-driven mutations that produces syntactically correct, shape-consistent, and trainable models, resolving the cold-start problem in learning-based NAS and allowing the language model to acquire architectural constraints before closed-loop refinement begins.

Experiments on CIFAR-100 show that the LLM can improve generated neural-network code in this setting: the best model achieves a 24.1\% relative improvement, increasing accuracy from 0.250 to 0.311 over the strongest model in the initial AST-generated distribution under the same proxy-evaluation protocol. Population-level and non-parametric analyses support an upward shift among valid candidates, while the discovered models reveal unconventional channel priors, including irregular non–power-of-two widths and late-stage expansion patterns. These results indicate that LLMs can be used as NAS optimizers for channel configuration search.

At the same time, the present study is limited to one dataset, one CNN backbone, and a channel-width search space, and it does not establish superiority over standard NAS baselines such as random search or evolutionary methods. The low validity rate also shows that unconstrained LLM generation remains inefficient under strict tensor-shape constraints. Because the framework operates on executable source code rather than a fixed supernet, it is naturally positioned for extension to broader architecture families and search spaces; validating this broader generality remains future work.
\vspace{0.4cm}

\noindent\textbf{Acknowledgments.}
This work was partially supported by the Alexander von Humboldt Foundation.

%
%
\bibliographystyle{splncs04}
\bibliography{references}

\end{document}